\documentclass{article}

\usepackage{microtype}
\usepackage{graphicx,multicol}
\usepackage{subfigure}
\usepackage{booktabs} 

\usepackage{times}
\usepackage{caption}
\usepackage{rotating,csquotes}
\usepackage{xcolor}
\usepackage{multirow}
\usepackage{framed}
\usepackage{booktabs}
\usepackage{amsmath}
\usepackage{upgreek}
\usepackage[normalem]{ulem}
\usepackage{amsfonts}
\usepackage{latexsym}
\usepackage{flexisym}
\usepackage{amsmath,graphicx}

\usepackage{hyperref}

\usepackage[accepted]{icml2021}

\begin{document}

\twocolumn[
\icmltitle{Transformer-based ASR Incorporating Time-reduction Layer and Fine-tuning with Self-Knowledge Distillation}





\begin{icmlauthorlist}
\icmlauthor{Md Akmal Haidar}{a}
\icmlauthor{Chao Xing}{a}
\icmlauthor{Mehdi Rezagholizadeh}{a}
\end{icmlauthorlist}

\icmlaffiliation{a}{Huawei Noah's Ark Lab, Montreal, Canada}

\icmlcorrespondingauthor{Md Akmal Haidar}{md.akmal.haidar@huawei.com}


\vskip 0.3in
]



\printAffiliationsAndNotice{}  

\begin{abstract}
End-to-end automatic speech recognition (ASR), unlike conventional ASR, does not have modules to learn the semantic representation from speech encoder. Moreover, the higher frame-rate of speech representation prevents the model to learn the semantic representation properly. Therefore, the models that are constructed by the lower frame-rate of speech encoder lead to better performance.  
For Transformer-based ASR, the lower frame-rate is not only important for learning better semantic representation but also for reducing the computational complexity due to the self-attention mechanism which has $O(n^2)$ order of complexity in both training and inference.
In this paper, we propose a Transformer-based ASR model with the time reduction layer, in which we incorporate time reduction layer inside transformer encoder layers in addition to traditional sub-sampling methods to input features that further reduce the frame-rate.
This can help in reducing the computational cost of the self-attention process for training and inference with performance improvement. Moreover, we introduce a fine-tuning approach for pre-trained ASR models using self-knowledge distillation (S-KD) which further improves the performance of our ASR model. Experiments on LibriSpeech datasets show that our proposed methods outperform all other Transformer-based ASR systems. 
Furthermore, with language model (LM) fusion, we achieve new state-of-the-art word error rate (WER) results for Transformer-based ASR models with just 30 million parameters trained without any external data.

\end{abstract}

\section{Introduction}

End-to-end (E2E) automatic speech recognition (ASR) systems~\citep{graves2006connectionist,amodei2016deep,chan2016listen,karita2019asru} have shown great success recently because of their simple training and inference procedures over the traditional HMM-based methods~\citep{povey}.  These end-to-end models learn a direct mapping of the input acoustic signal to the output transcription without needing to decompose the problem into different parts such as lexicon modeling, acoustic modeling and language modeling. The very first end-to-end ASR model is the connectionist temporal classification (CTC)~\citep{graves2006connectionist} model which independently maps the acoustic frames into the outputs. 
The CTC-based models which incorporate language model re-scoring could improve their output quality~\citep{graves2014towards}.
The conditional independence assumption in CTC was tackled by the recurrent neural network transducer (RNNT) model~\citep{graves2012sequence,He2018}, which shows better performance in streaming scenarios. Other E2E ASR models are the attention-based encoder-decoder (AED) architectures which yield state-of-the-art results in offline and online fashions~\citep{karita2019is, karita2019asru, chan2016listen, moritz2020, wang2020a,wang2020c, sony}. The Transformer architecture, which uses self-attention mechanism to model temporal contextual information, has been shown to achieve lower word error rate (WER) compared to the recurrent neural network (RNN) based AED architectures~\citep{karita2019asru}. 
However, Transformer architectures suffer from the decreased computational efficiency 
for longer input sequences because of quadratic time requirements of the self-attention mechanism. 

For ASR systems, the number of time frames for an audio input sequence is significantly higher than the number of output text labels. For example, a task such as LibriSpeech which has long utterances (15s), contains 1500 speech frames (assuming 10 ms per frame)~\citep{irie2019}. 
Considering that adjacent speech frames can form a chunk to represent more meaningful units like phonemes, some pre-processing mechanisms are considered to capture the embedding of a group of speech frames to reduce the frame-rate in the encoder input.  In~\citep{wang2020b}, different pre-processing strategies such as convolutional sub-sampling and frame stacking \& skipping techniques for Transformer-based ASR are discussed. Among these approaches, the convolutional approach for frame-rate reduction gives better WER results compared to other approaches. Recently, a VGG-like convolutional block~\citep{vgg2017} with max-pooling~\citep{hori2017} and layer normalization is used~\citep{wang2020a} before the Transformer encoder layers and outperforms the 2D-convolutional  sub-sampling~\citep{karita2019asru}. In~\citep{chan2016listen}, a pyramidal encoder structure for Bidirectional LSTM (BLSTM) is proposed using time-reduction layers to reduce the frame-rate of the input sequence by concatenating adjacent frames. In~\citep{He2018}, a time-reduction layer is employed in the RNNT model to speed up the training and inference. However, time-reduction layer for Transformer architecture was never explored.   

In this work, we hypothesize that further frame-rate reduction is possible on top of current approaches for Transformer-based architectures. In this regard, we introduce a time-reduction layer to Transformer-based ASR models that further decreases the frame-rate by concatenating adjacent frames.
Our proposed approach can speed up the Transformer model training and inference. Also, our model yields a new state-of-the art WER results for  Transformer-based ASR models 
over traditional approaches.  
Furthermore, we introduce a fine-tuning approach for pre-trained ASR  models incorporating the self-knowledge distillation (S-KD) approach~\citep{selfkdgeneralization,selfkdnlp}, which further improves the performance of our ASR model. S-KD is recently investigated which shows the dark knowledge of a model can be progressively deployed through distillation to improve even its own  performance~\citep{selfkdgeneralization,selfkdnlp}.
We summarize our contributions as following:
\begin{itemize}
    \item We introduce a Transformer-based ASR model incorporating  time-reduction layer;
    \item We deploy a fine-tuning approach for ASR model using S-KD which shows better performance than the conventional S-KD training from scratch. To best of our knowledge, we apply the S-KD approach for the first time in training ASR models;
    \item Our model outperforms the existing state-of-the-art Transformer-based ASR models. 
\end{itemize}

\section{Transformer Architecture for ASR}

In Transformer-based  ASR~\citep{karita2019asru, karita2019is}, the input sequence $X$ is first mapped to a subsequence $X_0 \in R^{n_{sub}\times d_{att}}$ by CNN blocks~\citep{dong2018speech,karita2019asru} and then transformer layers of the encoder map $X_0$ to a sequence of encoded features $X_e$. Here, $n_{sub}$ and $d_{att}$ describe the length of sub-sampled sequence and the dimensions of the features respectively. The layers of the encoder iteratively refine the representation of the input sequence with a combination of multi-head self-attention (MHA) and frame-level affine transformations. Specifically, the inputs to each layer are projected into keys $K$, queries $Q$, and values $V$. Scaled dot product attention is then used to compute a weighted sum of values for $Q$ matrix
~\citep{vaswani2017attention}:

\begin{equation}
\operatorname{Attention}(Q,K,V) = \operatorname{softmax}(\frac{QK^T}{\sqrt{d_{att}}})V
\end{equation}
where $Q \in R^{n_q \times d_{att}}$, $K,V \in R^{n_k\times d_{att}}$, $n_q$ is the length of $Q$, and $n_k$ is the length of $K,V$.
We obtain multi-head attention by performing this computation $h$ times independently with different sets of projections, and concatenating:

\begin{align} 
\operatorname{MHA}(Q,K,V) = \operatorname{Concat}(head_1,\dots,head_h)W^O \\
head_i = \operatorname{Attention}(QW_i^Q,KW_i^K,VW_i^V)
\end{align}
where $Q, K, V$ are inputs of this multi-head attention layer, $head_i \in R^{n_q \times d_{att}}$ is the $i^{th}$ attention layer output $(i=1,\dots , h)$, $W_i^* \in R^{d_{att}\times d_{att}}$, $W^O \in R^{d_{att}  h \times d_{att}}$ are the learnable weight matrices and $h$ is the number of attention heads in this layer~\citep{vaswani2017attention,karita2019asru}.
The outputs of multi-head attention  go through a 2-layer position-wise feed-forward network with hidden size $d_{ff}$:

\begin{equation} 
\operatorname{FFN}(X_0[t]) = \operatorname{ReLU}(X_0[t]W_1^{d_{ff}}+b_1^{d_{ff}})W_2^{d_{ff}} + b_2^{d_{ff}}
\label{eq:ffn}
\end{equation}
where $X_0[t] \in R^{d_{att}}$ 
is the $t^{th}$ frame of the sequence $X_0$, $W_1^{d_{ff}} \in R^{d_{att}\times d_{ff}}$, $W_2^{d_{ff}} \in R^{d_{ff}\times d_{att}}$ are learnable weight matrices, $b_1^{ff} \in R^{d_{ff}}$ and $b_2^{ff} \in R^{d_{att}}$ are learnable bias vectors. 

The decoder generates a transcription sequence $Y=(Y[1],...,Y[t])$ one token at a time. Each choice of output token $Y[t]$ is conditioned on the encoder representations $X_e$ and previously generated tokens $(Y[1],...,Y[t-1])$ through attention mechanisms. Each decoder layer performs two rounds of multi-head attention~\citep{karita2019asru}.
The multi-head attention is then followed by the position-wise FFN (Equation~\ref{eq:ffn}).  The output of the final decoder layer for token $Y[t-1]$ is used to predict the following token $Y[t]$. Other components of the architecture such as sinusoidal positional encodings, residual connections and layer normalization are described in~\citep{vaswani2017attention}. The positional encodings are applied into $X_0$ and $Y_0$ when convolutional sub-sampling was used~\citep{karita2019asru,espnet}. For VGG-like convolutional sub-sampling~\citep{wang2020a}, the positional encoding for $X_0$ was discarded.

\section{Related Work}
\label{sec:prior}

\textbf{Transformer ASR}
Several studies have focused on adapting Transformer networks for end-to-end speech recognition. In particular,~\citep{dong2018speech,mohamed2019transformers} present models augmenting Transformer networks with convolutions. \citep{karita2019asru} focuses on refining the training process, and show that Transformer-based end-to-end ASR is highly competitive with state-of-the-art methods. In~\citep{wang2020a}, a regularization method based on semantic masking was introduced for Transformer  ASR to force the decoder to learn a better language model. A hybrid Transformer model with deeper layers  and iterative loss was introduced in~\citep{wang2020b} where a Transformer-based acoustic model outperforms the best hybrid results  combined with an $n$-gram language model. In~\citep{e2e2020}, a semi-supervised model with pseudo-labeling using Transformer-based acoustic model was introduced. Recently a convolution augmented Transformer was proposed~\citep{conformer} where a convolution module with two macaron-like feed forward layers is augmented in the Transformer encoder layers. Moreover, Transformer architecture has shown better performance for streaming applications~\citep{transducer,moritz2020,sony}. In this paper, we incorporate  time-reduction layer in the vanilla Transformer architecture~\citep{vaswani2017attention, karita2019asru} and achieve better performance over the baselines~\citep{karita2019asru,wang2020a,wang2020b,moritz2020,e2e2020,specaugment}.       

\textbf{Knowledge Distillation} 
Knowledge distillation (KD) is a prominent neural model compression technique~\citep{hinton2015distilling} in which the output of a teacher network is used as an auxiliary supervision besides the ground-truth training labels. Later on, it was shown that KD can be used for improving the performance of neural networks in the so-called born-again~\citep{furlanello2018born} or self-distillation frameworks~\citep{selfkdgeneralization,yun2020regularizing,hahn2019self}. Self-distillation is a regularization technique trying to improve the performance of a network using its internal knowledge. In other words, in self-distillation scenarios, the student becomes its own teacher. 
While KD has shown great success in different ASR tasks~\citep{kdasra,kdasrb,kdasrc,kdasrd,kdasre,kdasrf,kdasrg}, self-distillation is more investigated in computer vision and natural language processing (NLP) domains~\citep{selfkdnlp,hahn2019self}. 
To best of our knowledge, we incorporate the self-KD approach for the first time in training ASR models.

\section{Proposed Model for Transformer ASR}
\begin{figure*}[ht]
\vskip 0.2in
\begin{center}
\centerline{\includegraphics[scale=1.0]{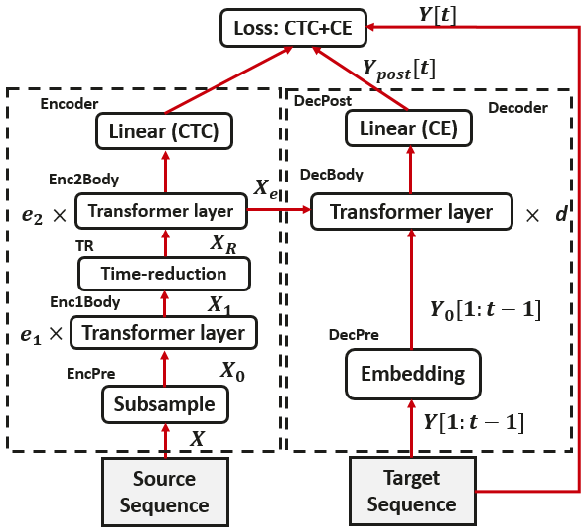}}
\caption{Proposed Transformer ASR with time-reduction layer}
\label{fig:TR}
\end{center}
\vskip -0.2in
\end{figure*}

Given an input sequence $X$ of log-mel filterbank speech features, Transformer predicts a target sequence $Y$ of characters or SentencePiece~\citep{kudo2018sentencepiece}. The architecture of our proposed model is described in Figure~\ref{fig:TR}.

\subsection{Transformer Encoder}
In Figure~\ref{fig:TR}, EncPre(.) transforms the source sequence $X$ into a sub-sampled sequence $X_0 \in R^{n_{sub}\times d_{att}}$ by using convolutional neural network (CNN) blocks~\citep{karita2019asru,moritz2020,wang2020a} which reduce the sequence length of the output $X_0$ by a factor of 4 compared to the sequence length of $X$. In the baseline model, the EncPre(.) is followed by a stack of Transformer layers that transform $X_0$ into a sequence of encoded features $X_e \in R^{n_{sub}\times d_{att}}$ for the CTC and decoder networks~\citep{karita2019asru}. In this paper, we introduce a time-reduction (TR) layer in the stack of Transformer layers of the encoder. The position of the TR layer in the Transformer layers is a  hyper-parameter. In Figure~\ref{fig:TR}, we add a TR layer between two groups (Enc1Body(.) and Enc2Body(.)) of Transformer layers in the encoder.
The Enc1Body(.) transforms $X_0$ into a sequence of encoded features $X_1 \in R^{n_{sub}\times d_{att}}$ for the TR layer. The TR layer transforms the sequence $X_1$ into $X_R \in R^{n_{sub2}\times d_{att}}$ by concatenating two adjacent time frames~\citep{chan2016listen}. This concatenation further reduces the length of the output sequence $X_R$ by a factor of 2 (i.e., the frame-rate of $X_R$ is reduced by a factor of 8 compared to $X$). 
Here, $n_{sub2}$ represents the sequence length after applying a TR layer. A TR layer corresponding to the $j^{th}$ layer at the output of $i^{th}$ time step can be described as follows:
\begin{equation}
h_i^j = \left[h_{2i}^{j-1}, h_{2i+1}^{j-1}\right]
\label{eq:tr}
\end{equation} 
where $h_i^j$ represents the encoder representation of the $j^{th}$ layer at the $i^{th}$ time step. After the TR layer, Enc2Body(.) transforms $X_R$ into a sequence of encoded features $X_e \in R^{n_{sub2}\times d_{att}}$ for the CTC and decoder networks. The Enc1Body(.) and Enc2Body(.) can be defined as~\citep{karita2019asru}:
\begin{flalign}
\begin{split}
X_i^\textprime &= X_i+MHA_i(X_i, X_i, X_i)\\
X_{i+1}&=FFN_i(X_i^\textprime)
\end{split}
\label{eq:mha}
\end{flalign}
where $i=0,\dots,e_1-1$ or $i=0,\dots,e_2-1$ describe the index of the Transformer layers of the encoder before or after the TR layer respectively.

Since the time complexity of the self-attention mechanism is quadratic, our proposed time-reduction layer inside the Transformer layers of the encoder can reduce the computational cost for training and inference. To best of our knowledge, we are the first who applied time-reduction approach to Transformers.  


\subsection{Transformer Decoder}
We keep the same decoder architecture as in~\citep{karita2019asru, karita2019is}. The decoder receives the encoded sequence $X_e$ and the prefix of a target sequence $Y[1:t-1]$ of token IDs: characters or SentencePiece~\citep{kudo2018sentencepiece}. First, DecPre(.) embeds the tokens into learnable vectors $Y_0[1:t-1]$. Then, DecBody(.) and a single linear layer DecPost(.) predicts the posterior distribution of the next token prediction $Y_{post}[t]$ given $X_e$ and $Y[1:t-1]$. For the decoder input $Y[1:t-1]$, we use ground-truth labels in the training stage, while we use generated output in the decoding stage. The DecBody(.) can be described as:

\begin{flalign}
\begin{split}
     Y_j^\textprime[t] &= Y_j[t]+MHA_j^{self}(Y_j[t],Y_j[1:t-1], Y_j[1:t-1])\\
    Y_j^{\textprime\textprime}&=Y_j+MHA_j^{src}(Y_j^\textprime,X_e,X_e)\\
    Y_{j+1}&=Y_j^{\textprime\textprime}+FFN_j(Y_j^{\textprime\textprime})
\end{split}
\end{flalign}
where $j=0,\dots,d-1$ represents the index of the Transformer layers of the decoder. $MHA_j^{src}(Y_j^\textprime,X_e,X_e)$ and $MHA_j^{self}(Y_j[t],Y_j[1:t-1], Y_j[1:t-1])$ are defined as the  'encoder-decoder attention' and the 'decoder self-attention' respectively. 

\subsection{Training and Decoding}
During ASR training, the frame-wise posterior distribution of $P_{s2s}(Y|X)$ and $P_{ctc}(Y|X)$ are predicted by the decoder and the CTC module respectively. The training loss function is the weighted sum of the negative log-likelihood of these values~\citep{karita2019asru}:

\begin{equation}
L_{ASR}=-(1-\alpha) \log P_{s2s}(Y|X)-\alpha \log P_{ctc}(Y|X)    
\end{equation}

where$-\log P_{ctc}(Y|X)$ and $-\log P_{s2s}(Y|X)$ describe the CTC and cross-entropy (CE) losses respectively. $\alpha$ is a hyper-parameter which determines the CTC weight.    
In the decoding stage, given the speech feature $X$ and the previous predicted token, the next token is predicted using beam search combining the scores of s2s and ctc with/without language model score as~\citep{karita2019asru}:

\begin{align}
\begin{split}
\hat{Y}=\underset{Y \in y^*}{\arg\max}\{(1-\lambda) \log P_{s2s}(Y|X_e)+ \\ \lambda \log P_{ctc}(Y|X_e) + \gamma \log P_{lm}(Y)\}  
\end{split}
\end{align}
where $y^*$is a set of hypotheses of the target sequence $Y$, and $\lambda$, $\gamma$ are hyperparameters.

\begin{figure*}[ht]
\vskip 0.2in
    \begin{center}
    \centerline{\includegraphics[scale=0.8]{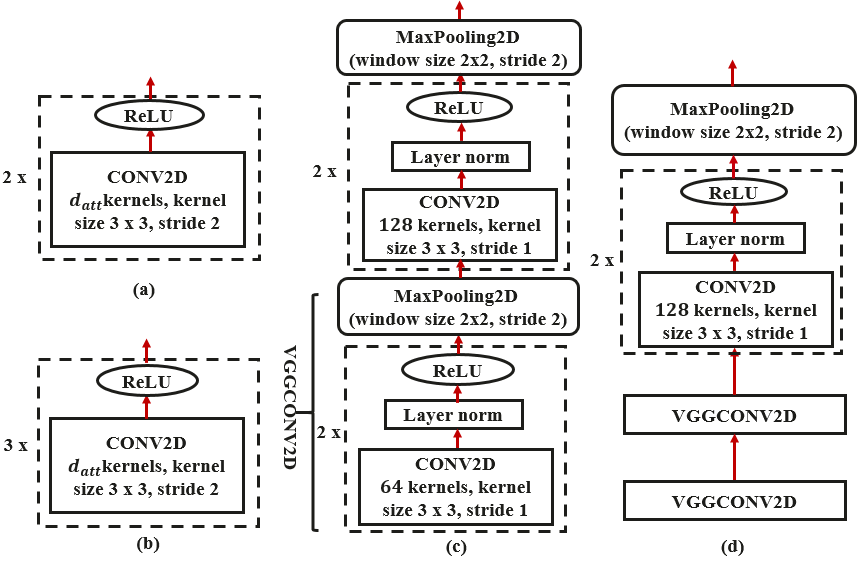}}
    \caption{Different types of Encoder sub-sampling. (a) CONV2D4, (b) CONV2D8, (c) VGGCONV2D4, and (d) VGGCONV2D8}
    \label{fig:encpre}
    \end{center}
\vskip -0.2in
\end{figure*}
\subsection{Encoder sub-sampling}

We apply 2D-convolutional sub-sampling~\citep{karita2019asru} (CONV2D4) for EncPre(.) which is a block of two CNN layers. Each CNN layer uses stride 2, a kernel size of $3\times3$ and a ReLU activation function. For EncPre(.), we also use VGG-like convolutional block (VGGCONV2D4) with layer normalization and a max-pooling function~\citep{wang2020a}. For both cases, the sequence length reduces by a factor of 4 compared to the length of $X$. To compare with our proposed approach, we also define two other EncPre(.) namely CONV2D8~\citep{espnet} and VGGCONV2D8 which reduce the sequence length by 8 times compared to the original sequence length. 
All the sub-sampling approaches are depicted in Figure~\ref{fig:encpre}. 

\section{Fine-tuning using Self-Knowledge Distillation}
\begin{figure*}[ht]
\vskip 0.2in
\begin{center}
    
    \centerline{\includegraphics[scale=0.85]{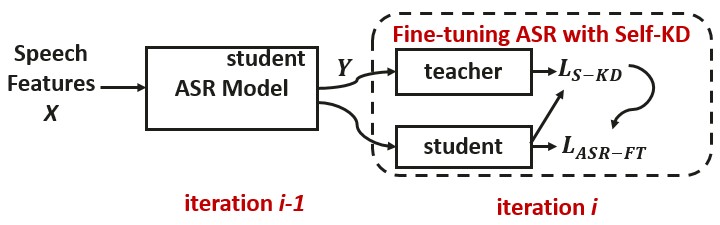}}
    \caption{Fine-tuning ASR using the S-KD approch. The student at $i-1$ becomes the teacher at iteration $i$ and the model is trained with the loss $L_{ASR-FT}$}
    \label{fig:s-kd}
    \end{center}
    \vskip -0.2in

\end{figure*}

We further improve the performance of our proposed model by incorporating a self-knowledge distillation (S-KD) approach~\citep{selfkdgeneralization}. S-KD has recently been investigated for natural language processing (NLP) applications, where the soft labels generated by the model are used to improve itself in a progressive manner~\citep{selfkdnlp, selfkdgeneralization}. S-KD is a framework of KD where the student becomes its own teacher. Bear in mind, since ASR training takes longer time to converge, starting from scratch with S-KD as in~\citep{selfkdgeneralization} would increase the training time significantly. In this paper, we incorporate S-KD in fine-tuning a pre-trained ASR model and will show that it will be more efficient than training ASR models using S-KD from scratch. For fine-tuning using S-KD, let $P^T_{s2s}(Y[t]|Y[1:t-1], X_e)$ be the probability distribution of the decoder of a pre-trained ASR model as a teacher. We initialized the student with the teacher model and at iteration $i-1$, the student model tries to mimic the teacher probability distribution with cross-entropy:

\begin{align}
    \begin{split}
L_{S-KD}=-\sum_{Y[t] \in V} (P^T_{s2s}(Y[t]|Y[1:t-1], X_e) \times\\
\log P_{s2s}(Y[t]|Y[1:t-1], X_e))
    \end{split}
\end{align}
where $V$ is a set of vocabulary. At iteration $i$, the student probability distribution $P_{s2s}(Y[t]|Y[1:t-1], X_e)$ of the decoder becomes the teacher probability distribution $P^T_{s2s}(Y[t]|Y[1:t-1],X_e)$ and the student model is trained to mimic the teacher with $L_{S-KD}$.
A schematic of the S-KD approach is described in Figure~\ref{fig:s-kd}. 
The new loss function for fine-tuning (FT) the ASR model can be described as:    

\begin{align}
\begin{split}
L_{ASR-FT}=\alpha L_{ctc} +(1-\alpha) (\phi_{KD} L_{S-KD} \\ +(1-\phi_{KD})L_{s2s})
\end{split}
\end{align}
where $L_{ctc}=-\log P_{ctc}(Y|X)$, $L_{s2s}=-\log P_{s2s}(Y|X)$, and $\phi_{KD}$ is a controlled parameter, and is typically $\phi_{KD}=0.5$~\citep{sony}. 

To show the effectiveness of our proposed fine-tuning approach using  S-KD, we also perform S-KD experiments to train our model from scratch. In this regard, we update the KD loss for each epoch similar to~\citep{selfkdgeneralization}:
\begin{equation}
\begin{split}
   & \phi_{KD_t} = \phi_{KD_T} \times \frac{t}{T} \\
\end{split}
\end{equation}
where $T$ is the total number of training epoch\sout{s}, $\phi_{KD_T}$ is the $\phi_{KD_t}$ for the last epoch. 

\section{Experiments}
\label{sec:experiments}
\subsection{Experimental Setup}
We use the open-source, ESPnet toolkit~\citep{espnet} for our experiments. We conduct our experiments on LibriSpeech dataset~\citep{panayotov2015librispeech}, which is a speech corpus of reading English audio books. It has 960 hours of training data, 10.7 hours of development data, and 10.5 hours of test data, whereby the development and the test datasets are both split into approximately two halves named “clean” and “other”. To extract the input features for speech, we follow the same setup as in~\citep{espnet, karita2019asru}: using 83-dimensional log-Mel filterbanks frames with pitch features~\citep{karita2019asru,moritz2020}. The output tokens come from a 5K subword vocabulary created with sentencepiece~\citep{kudo2018sentencepiece}``unigram''~\citep{espnet}. 
We perform experiments using 12 ($e_1+e_2=12$) and 6 ($d=6$) Transformer layers for the encoder and decoder respectively, with $d_{ff}=2048$, $d_{att}=256$, and $h=4$. The total number of model parameters are about 30 M.
We apply default settings of SpecAugmentation~\citep{specaugment} of ESPnet~\citep{espnet} and no gradient accumulation for all of our experiments. We use Adam optimizer with learning rate scheduling similar to~\citep{vaswani2017attention,espnet} and other  settings (e.g., dropout, warmup steps, $\alpha$, label smoothing penalty~\citep{specaugment}) following~\citep{moritz2020,espnet}. We set the initial learning rate of 5.0~\citep{espnet}. We train all of our models for 150 epochs. We report our results based on averaging the best five checkpoints. 
For fine-tuning with self-knowledge distillation (FS-KD) experiments, we use our trained best average model for initialization. We train the FS-KD experiments for 50 epochs with Adam optimizer~\citep{adam} with fixed learning rate of 0.0001~\citep{espnet} and fixed $\phi_{KD}=0.5$. We also perform S-KD experiments from scratch for 150 epochs with $\phi_{KD_T}=0.5$ and 200 epochs with $\phi_{KD_T}=0.7$ respectively. For S-KD experiments, We use Adam optimizer with learning rate scheduling similar to~\citep{vaswani2017attention,espnet}.

For decoding, the CTC weight $\lambda$, the LM weight $\gamma$ and beam size are 0.5, 0.7 and 20 respectively~\citep{moritz2020,espnet}. Moreover, we apply insertion-penalty of 2.0 during decoding. For S-KD and FS-KD experiments, we note that the CTC weight of $\lambda=0.3$ gives better results. This might be because of S-KD applied on the decoder output of the ASR model. For LM fusion~\citep{shallow2018}, we use a pre-trained Transformer LM\footnote{ \url{https://github.com/espnet/espnet/blob/master/egs/librispeech/asr1/RESULTS.md}}.

\begin{table*}[ht]
\caption{WER results for E2E ASR models. TR0 and TR2 represent the time-reduction (TR) layer applied before all and after second Transformer layers of the encoder respectively. Best results are depicted in bold.}
\label{tab:results}
\vskip 0.15in
\begin{center}
\begin{small}
\begin{sc}
\begin{tabular}{lcccc}
\toprule
    Model       &Test-clean&Test-other&Dev-clean&Dev-other \\
\midrule
    Hybrid Transformer+LM~\citep{wang2020b} &2.26&4.85&-&- \\
    RWTH+LM~\citep{rwth}&2.3&5.0&1.9&4.5\\
    Baseline +LM ~\citep{moritz2020} &2.7  &6.1   &2.4  &6.0\\
    ESPNet Transformer+LM~\citep{karita2019asru}& 2.6  &5.7   & 2.2 &5.6\\
    LAS+LM~\citep{specaugment}&3.2&9.8&-&-\\
    LAS+SpecAugment+LM~\citep{specaugment} &2.5 &5.8 &- &- \\
    Transformer+LM~\citep{e2e2020}&2.33&5.17&2.10&4.79 \\
    Semantic Mask + LM ~\citep{wang2020a}&2.24 &5.12 &2.05 &5.01 \\
    Transformer Transducer+LM~\citep{transducer}&2.0&4.6&-&- \\
    \hline
        New Baselines &&&& \\ \hline
    Pyramidal Transformer  & 4.3 &9.3  &4.2 &9.1 \\ 
     \hspace{0.5cm}  + LM &2.9 &5.9 &2.7 &5.7 \\
    CONV2D8  & 4.1&9.1 & 3.9&9.5\\
     \hspace{0.5cm}  + LM &2.7 &5.6 &2.4 &5.6 \\
    VGGCONV2D8   &3.6 & 9.2  & 3.5 & 9.4\\
     \hspace{0.5cm}  + LM & 2.0& 5.3&1.9 &5.2 \\

\hline
    Proposed Models with TR &&&& \\ \hline
    CONV2D4+TR0  &3.6 & 8.6 &3.4  & 8.9 \\
    \hspace{0.5cm}   + LM &2.3 &5.3 &2.0 &5.0\\
    VGGCONV2D4+TR0 & 3.5&8.5 & 3.3 & 8.8 \\
    \hspace{0.5cm}   + LM &\textbf{2.0} &\textbf{5.0} &\textbf{1.9} &\textbf{4.9}\\
    CONV2D4+TR2 &3.7 & 8.2 & 3.5 & 8.5 \\
    \hspace{0.5cm}   + LM &2.4 &5.2 &2.2 &5.0\\
    VGGCONV2D4+TR2 &3.3 & 8.5 & 3.2 & 8.5 \\
    \hspace{0.5cm}   + LM &\textbf{2.0} &\textbf{5.0} &\textbf{1.9} &\textbf{4.9}\\
\bottomrule
\end{tabular}
\end{sc}
\end{small}
\end{center}
\vskip -0.1in
\end{table*}

\subsection{New Baseline Results}
We compare our proposed models with  state-of-the-art models~\citep{moritz2020,specaugment,karita2019asru,wang2020a,wang2020b,rwth,e2e2020}. We also introduce the latest baseline models and run experiments to compare them with our proposed approaches. We create a pyramidal structure~\citep{chan2016listen} in the first three Transformer layers of the encoder, where we concatenate the outputs at consecutive steps (Equation~\ref{eq:tr}) of each layer before feeding it to the next layer. After the first three layers, the sequence length would reduce by a factor of 8 
than the sequence length of $X$. We define the model as Pyramidal Transformer. We also use CONV2D8 sub-sampling~\citep{espnet} and VGGCONV2D8 which are explained in Figure~\ref{fig:encpre} for EncPre(.) to reduce the sequence length 8 times than the length of $X$ and then apply all the Transformer layers. We define them as  CONV2D8 and VGGCONV2D8 respectively. 
From Table~\ref{tab:results}, we can see that our proposed new baseline models give comparable results to the existing models with LM fusion, which are much larger than our model. Among these baselines, the VGGCONV2D8 gives the better results.

\subsection{Results using TR layer}
We apply the TR layer after CONV2D4 or VGGCONV2D4 to further reduce the sequence length by a factor of two. 
First, we apply the TR layer after CONV2D4 and VGGCONV2D4 approaches (i.e., TR0: $e_1=0$ and $e_2=12$). We define the models as CONV2D4+TR0 and VGGCONV2D4+TR0 respectively, which reduce the frame-rate by a factor of 8 than the frame-rate of $X$.
From Table~\ref{tab:results}, 
we can note that VGGCONV2D4+TR0 outperforms or gives comparable results to the existing hybrid results~\citep{wang2020c,rwth}, E2E LSTM~\citep{specaugment} and E2E Transformer baselines~\citep{moritz2020,karita2019asru,e2e2020,wang2020a} which are trained with more parameters. Also, it shows better results over the VGGCONV2D8 model. 
Next, we apply the TR layer after the second Transformer layer (i.e., TR2: $e_1=2$ and $e_2=10$).  From both TR0 and TR2 experiments, we can see that both experiments give comparable results and VGGCONV2D4+TR0/TR2 yields better results over the CONV2D4+TR0/TR2 models. Without LM fusion, VGGCONV2D4+TR2 gives better results over VGGCONV2D4+TR0. 
For the other experiments, we apply the VGGCONV2D4 for the EncPre(.) and perform TR2 experiments.
From the table~\ref{tab:results}, we see that the VGGCONV2D4+TR0/TR2 model with LM fusion outperforms over the best E2E Transformer ASR models  using the semantic mask~\citep{wang2020a} and~\citep{e2e2020} which are  trained with 75 and 270 M parameters respectively. Compared to the semantic mask technique~\citep{wang2020a} and ~\citep{e2e2020}, our proposed approach achieves relative WER reductions of 10.7\% and 14.1\% for the test-clean and 2.3\% and 3.2\% for the test-other respectively. 
Also, we can note that our approach shows comparable results to Transformer Transducer~\citep{transducer} which is trained with RNN-T loss~\citep{rnnt-loss} and 139 M parameters. 
\begin{table*}[ht]
\caption{WER results of the proposed Transformer-based ASR models using self-KD. FS-KD describes the results for fine-tuning the pre-trained VGGCONV2D4+TR2 model with Self-KD. TR2\_S-KD* and TR2\_S-KD** represent the models training from scratch using self-KD with   time-reduction layer after second Transformer layer for 150 and 200 epochs respectively. Best results are described in bold.}
\label{tab:s-kdresults}
\vskip 0.1in
\begin{center}
\begin{small}
\begin{sc}
\begin{tabular}{lcccc}
\toprule
    Model       &Test-clean&Test-other&Dev-clean&Dev-other \\
\midrule
     VGGCONV2d4+TR2 +FS-KD& 3.1& 7.9 &3.0  & 8.0 \\
    \hspace{0.5cm}   + LM &\textbf{1.9} &\textbf{4.8} &\textbf{1.8} &\textbf{4.6}\\
     VGGCONV2d4+TR2\_S-KD* & 3.2& 8.1 &3.2  & 8.3 \\
    \hspace{0.5cm}   + LM & 2.0&4.8 &1.9 &4.7\\
     VGGCONV2d4+TR2\_S-KD*+FS-KD & 3.3& 8.1 &3.2  & 8.3 \\
    \hspace{0.5cm}   + LM &2.0 &4.8 &1.8 &4.7\\ 
    VGGCONV2d4+TR2\_S-KD**  & 3.2& 8.0 &3.1  & 8.2 \\
    \hspace{0.5cm}   + LM &2.0 &4.9 & 1.9&4.8\\
\bottomrule
\end{tabular}
\end{sc}
\end{small}
\end{center}
\vskip -0.1in
\end{table*}

\subsection{Results using Self-KD}
We report the results for self-KD (S-KD) experiments in Table~\ref{tab:s-kdresults}. We conduct experiments VGGCONV2D4+TR2+FS-KD for fine-tuning the pre-trained ASR model VGGCONV2D4+TR2 using S-KD. We can note that with FS-KD, the proposed method gives further WER reductions for both without and with LM fusion. With LM fusion, our model outperforms the Transformer transducer model for the test-clean dataset and gives comparable results for the test-other dataset. We also train an ASR model by applying S-KD from scratch to show the effectiveness of our fine-tuning approach on a pre-trained ASR model. We conduct two experiments by applying self-KD from scratch for 150 epochs and 200 epochs respectively. From Table~\ref{tab:s-kdresults}, we can see that the S-KD experiments from scratch give similar or better results than the VGGCONV2D4+TR2 model but lower performance than our proposed VGGCONV2D4+FS-KD approach. Also, the self-KD approach from scratch requires longer training time than our proposed approaches. Furthermore, to make the same number of epochs, we apply self-KD for fine-tuning the VGGCONV2D4+TR2\_S-KD* model which is trained using the self-KD from scratch for 150 epochs. From Table~\ref{tab:s-kdresults}, it can be seen that the model VGGCONV2D4+TR2\_S-KD*+FS-KD  does not give better results than our proposed model VGGCONV2D4+TR2+FS-KD.




      

\begin{table}[ht]
\caption{Rows 1-5 represent the best five validation accuracy for the VGGCONV2D4+TR2, VGGCONV2D4+TR2+FS-KD, VGGCONV2D4+TR2+S-KD*, VGGCONV2D4+TR2_S-KD*+FS-KD, and VGGCONV2D4+TR2_S-KD**. Best results are in bold.}
\label{val5}
\vskip 0.1in
\begin{center}
\begin{small}
\begin{sc}
\begin{tabular}{ccccc}
\toprule
    1&2&3&4&5 \\
\midrule
    0.95053& 0.94900& 0.94793& 0.94780& 0.94735 \\ 
    \textbf{0.95518}& \textbf{0.95475}& \textbf{0.95436}& \textbf{0.95434}& \textbf{0.95418} \\ 
    0.95043 & 0.95011& 0.95002& 0.94999& 0.94966 \\ 
    0.95209& 0.95184& 0.95072& 0.95049& 0.94973 \\ 
    0.95266& 0.95216& 0.95189& 0.95187& 0.95146 \\ 
\bottomrule
\end{tabular}
\end{sc}
\end{small}
\end{center}
\vskip -0.1in
\end{table}



Moreover, we  report the  best five  validation  accuracy for our best model with the time-reduction layer and the models with self-KD in Table 2. We observe that our proposed fine-tuning approach using self-KD (VGGCONV2D4+TR2+FS-KD) gives better validation accuracy over the other models. 

\subsection{Time Complexity Analysis}
Transformers are recently the most efficient models in speech and semantic fields which were suffered from a $(1+\frac{m}{l})^2$ decay in computational efficiency for the self-attention mechanism if the sequence length $l$ is increased by $m$ steps. Our proposed approach could release the burden of computational efficiency in Transformers by applying time reduction layer. For example, if the sequence length $l$ is decreased by a factor of 2, the computation for self-attention can be reduced by $(\frac{l}{\frac{l}{2}})^2$ times. Therefore, the proposed method can reduce the computational cost of the self-attention mechanism by $k^2$ times over the traditional approach, where $k$ is the time reduction ratio. 


\section{Conclusion and Future Work}
\label{sec:conclusion}
Convolutional sub-sampling approaches are essential for efficient use of self-attention mechanism in Transformer-based ASR. These sub-sampling approaches reduce the speech frame-rate to form meaningful units like phoneme, word-piece, etc. The reduced frame-rate can help in proper training of Transformer ASR. In this paper, we incorporated a time-reduction layer for Transformer-based ASR which can further reduce the frame-rate and improve the performance over traditional convolutional sub-sampling approaches. It can also gives faster training and inference during the self-attention computation of the Transformer ASR model. Our approach yields state-of-the-art results using the LibriSpeech benchmark dataset for Transformer  architectures. Furthermore, we introduced fine-tuning of a pre-trained ASR model using self-knowledge distillation which further improves the performance of our ASR model. Moreover, we showed our proposed fine-tuning approach with self-KD outperformed the conventional self-KD training from scratch. We performed experiments on LibriSpeech datasets and show that our proposed method with 30 M parameters achieves word error rate (WER) of \textbf{1.9\%}, \textbf{4.8\%}, \textbf{1.8\%} and \textbf{4.6\%} results for the test-clean, test-other, dev-clean and dev-other respectively using language model (LM) fusion and outperformed over the other Transformer-based ASR models. For future work, we will investigate our approaches for conformer~\citep{conformer} architecture and explore more experimental settings with incorporating gradient accumulation.


\bibliography{example_paper}
\bibliographystyle{icml2021}


%



\end{document}